\documentclass[10pt,twocolumn,letterpaper]{article}

\usepackage{cvpr}
\usepackage{times}
\usepackage{epsfig}
\usepackage{graphicx}
\usepackage{amsmath}
\usepackage{amssymb}

\usepackage{booktabs}

\usepackage[pagebackref=true,breaklinks=true,letterpaper=true,colorlinks,bookmarks=false]{hyperref}

\cvprfinalcopy %

\def\cvprPaperID{1916} %

\ifcvprfinal\fi
\begin{document}

\title{CRNet: Cross-Reference Networks for Few-Shot Segmentation}

\author{Weide Liu$^1$, ~~ Chi Zhang$^1$, ~~ Guosheng Lin$^1$\thanks{Corresponding author: G. Lin (e-mail: {\tt gslin@ntu.edu.sg})}, ~~ Fayao Liu$^2$\\
	$^{1}$Nanyang Technological University, Singapore\\
	$^{2}$A*Star, Singapore\\
E-mail: {$\tt weide001@e.ntu.edu.sg$}, { $\tt chi007@e.ntu.edu.sg$ }, { $\tt gslin@ntu.edu.sg$ }
}

\maketitle
\thispagestyle{empty}

\begin{abstract}
Over the past few years, state-of-the-art image segmentation algorithms are based on deep convolutional neural networks. To render a deep network with the ability to understand a concept, humans need to collect a large amount of pixel-level annotated data to train the models, which is time-consuming and tedious. Recently, few-shot segmentation is proposed to solve this problem. Few-shot segmentation aims to learn a segmentation model that can be generalized to novel classes with only a few training images. 
In this paper, we propose a cross-reference network (CRNet) for few-shot segmentation. Unlike previous works which only predict the mask in the query image, our proposed model concurrently make predictions for both the support image and the query image. With a cross-reference mechanism, our network can better find the co-occurrent objects in the two images, thus helping the few-shot segmentation task. We also develop a mask refinement module to recurrently refine the prediction of the foreground regions. For the $k$-shot learning, we propose to finetune parts of networks to take advantage of multiple labeled support images. Experiments on the PASCAL VOC 2012 dataset show that our network achieves state-of-the-art performance.

\end{abstract}
\section{Introduction}

Deep neural networks have been widely applied to visual understanding tasks, \eg, objection detection, semantic segmentation and image captioning, since the huge success in ImageNet classification challenge~\cite{imagenet}. Due to its data-driving property, large-scale labeled datasets are often required to enable the training of deep models. 
However, collecting labeled data can be notoriously expensive in tasks like semantic segmentation, instance segmentation, and video segmentation. Moreover, data collecting is usually for a set of specific categories. Knowledge learned in previous classes can hardly be transferred to unseen classes directly.
Directly finetuning the trained models still needs a large amount of new labeled data. Few-shot learning, on the other hand, is proposed to solve this problem. In the few-shot learning tasks, models trained on previous tasks are expected to generalize to unseen tasks with only a few labeled training images.

In this paper, we target at few-shot image segmentation. Given a novel object category, few-shot segmentation aims to find the foreground regions of this category only seeing a few labeled examples. Many previous works formulate the few-shot segmentation task as a guided segmentation task. The guidance information is extracted from the labeled support set for the foreground prediction in the query image, which is usually achieved by an unsymmetrical two-branch network structure. The model is optimized with the ground truth query mask as the supervision.
 
 \begin{figure}[t]
    \includegraphics[width=1\linewidth]{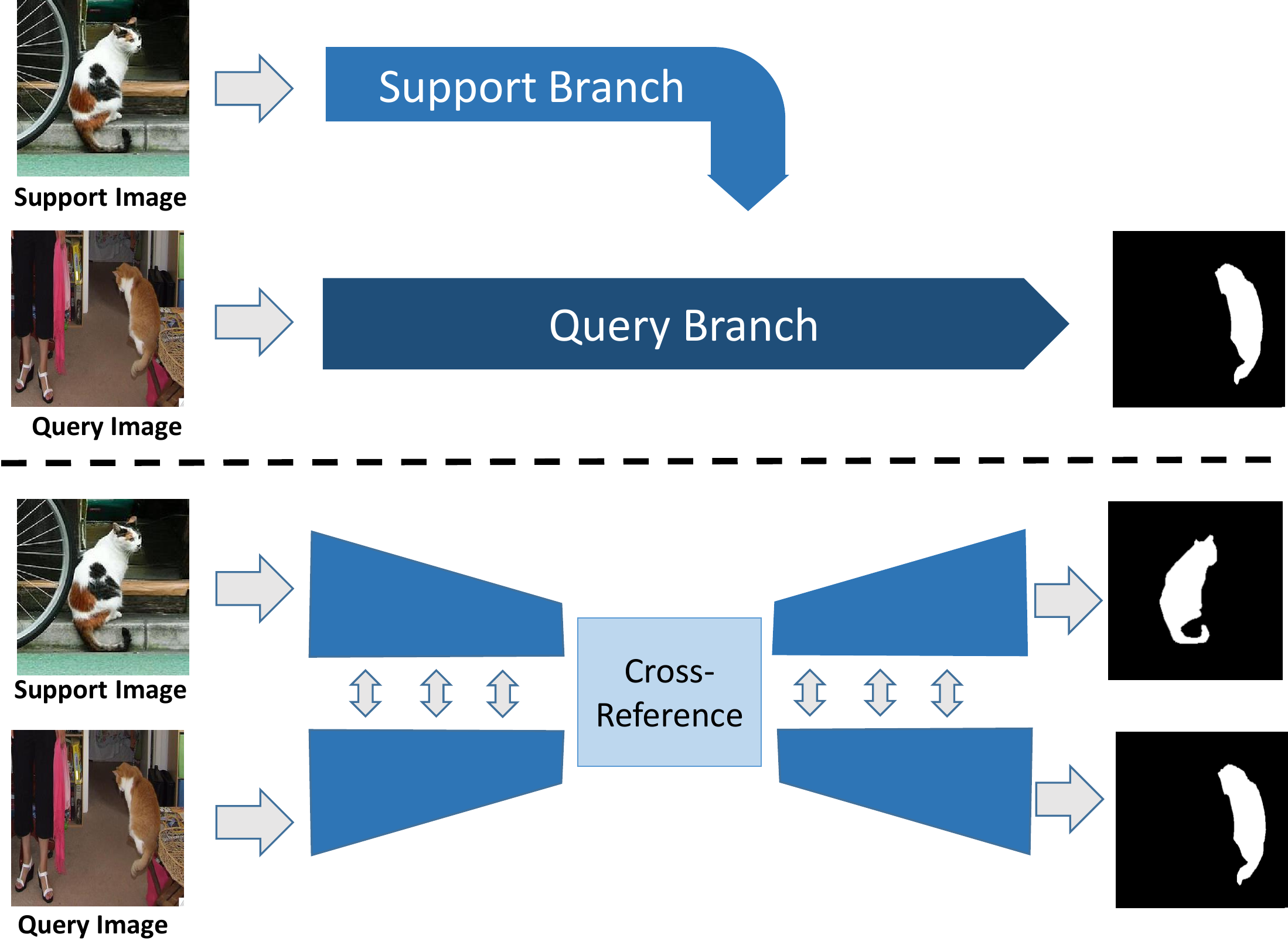}
    \caption{Comparison of our proposed CRNet against previous work. Previous work (upper part) unilaterally guide the segmentation of query images with support images, while in our CRNet (lower part) support and query images can guide the segmentation of each other. }
    \label{twopair}
        \vskip +1em
\end{figure}

In our work, we argue that the roles of query and support sets can be switched in a few-shot segmentation model. Specifically, the support images can guide the prediction of the query set, and conversely, the query image can also help make predictions of the support set. Inspired by the image co-segmentation literature~\cite{joulin2012multi,mukherjee2018object,chen2018semantic}, we propose a symmetric Cross-Reference Network that two heads concurrently make predictions for both the query image and the support image.
The difference of the network design with previous works is shown in Fig.~\ref{twopair}.
The key component in our network design is the cross-reference module which generates the reinforced feature representations by comparing the co-occurrent features in two images. The reinforced representations are used for the downstream foreground predictions in two images. In the meantime, the cross-reference module also makes predictions of co-occurrent objects in the two images.  This sub-task provides an auxiliary loss in the training phase to facilitate the training of the cross-reference module. 

As there exists huge variance in the object appearance, mining foreground regions in images can be a multi-step process. We develop an effective Mask Refinement Module to iteratively refine our predictions. 
In the initial prediction, the network is expected to locate high-confidence seed regions. Then, the confidence map, in the form of probability map, is saved as the cache in the module and is used for later predictions. We update the cache every time we make a new prediction. After running the mask refinement module for a few steps, our model can better predict the foreground regions. We empirically demonstrate that such a light-weight module can significantly improve the performance.

When it comes to the $k$-shot image segmentation where more than one support images are provided, previous methods often use 1-shot model to make predictions with each support image individually and fuse their features or predicted masks. 
In our paper, we propose to finetune parts of our network with the labeled support examples. As our network can make predictions for both two image inputs at a time, we can use at most $k^2$ image pairs to finetune our network. An advantage of our finetuning based method is that it can benefit from the increasing number of support images, and thus consistently increases the accuracy. In comparison, the fusion-based methods can easily saturate when more support images are provided. In our experiment, we validate our model in the 1-shot, 5-shot, and 10-shot settings.

The main contributions of this paper are listed as follows: 

\begin{itemize}
\item We propose a novel cross-reference network that concurrently makes predictions for both the query set and the support set in the few-shot image segmentation task. By mining the co-occurrent features in two images, our proposed network can effectively improve the results.
\item We develop a mask refinement module with confidence cache that is able to recurrently refine the predicted results.
\item We propose a finetuning scheme for $k$-shot learning, which turns out to be an effective solution to handle multiple support images. 
\item Experiments on the PASCAL VOC 2012 demonstrate that our method significantly outperforms baseline results and achieves new state-of-the-art performance on the 5-shot segmentation task. 

\end{itemize}

\section{Related Work}
\subsection{Few shot learning}
Few-shot learning aims to learn a model which can be easily transferred to new tasks with limited training data available. Few-shot learning is widely explored in image classification tasks. Previous methods can be roughly divided into two categories based on whether the model needs finetuning at the testing time. In non-finetuned methods, parameters learned at the training time are kept fixed at the testing stage. For example, ~\cite{snell2017prototypical,relation,vinyals2016matching,zhangcvpr} are metric based approaches where an embedding encoder and a distance metric are learned to determine the image pair similarity. These methods have the advantage of fast inference without further parameter adaptions. However, when multiple support images are available, the performance can become saturate easily. In finetuning based methods, the model parameters need to be adapted to the new tasks for predictions. For example,  in~\cite{chen2019closerfewshot}, they demonstrate that by only finetuning the fully connected layer, models learned on training classes can yield state-of-the-art few-shot performance on new classes. In our work, we use a non-finetuned feed-forward model to handle 1-shot learning and adopt model finetuning in the $k$-shot setting to benefit from multiple labeled support images. The task of few-shot learning is also related to open set problem~\cite{sun2020conditional}, where the goal is only to detect data from novel classes.

\subsection{Segmentation}
Semantic segmentation is a fundamental computer vision task which aims to classify each pixel in the image. State-of-the-art methods formulate image segmentation as a dense prediction task and adopt fully convolutional networks to make predictions~\cite{chen2018deeplab,long2015fully}. Usually, a pre-trained classification network is used as the network backbone by removing the fully connected layers at the end. To make pixel-level dense predictions, encoder-decoder structures~\cite{lin2017refinenet,long2015fully} are often used to reconstruct high-resolution prediction maps. Typically an encoder gradually downsamples the feature maps, which aims to acquire large field-of-view and capture abstract feature representations. Then, the decoder gradually recovers the fine-grained information. Skip connections are often used to fuse high-level and low-level features for better predictions. In our network, we also follow the encoder-decoder design and opt to transfer the guidance information in the low-resolution maps and use decoders to recover details.

\subsection{Few-shot segmentation}
Few-shot segmentation is a natural extension of few-shot classification to pixel levels. Since Shaban~\etal~\cite{shaban2017one} propose this task for the first time, many deep learning-based methods are proposed. Most previous works formulate the few-shot segmentation as a guided segmentation task. For example, in~\cite{shaban2017one},  the side branch takes the labeled support image as the input and regress the network parameters in the main branch to make foreground predictions for the query image. In~\cite{zhang2019canet}, they share the same spirits and propose to fuse the embeddings of the support branches into the query branch with a dense comparison module.
Dong~\etal~\cite{Dong2018FewShotSS} draw inspiration from the success of Prototypical Network~\cite{snell2017prototypical} in few-shot classifications, and propose a dense prototype learning with Euclidean distance as the metric for segmentation tasks. Similarly, Zhang~\etal~\cite{zhang2018sg} propose a cosine similarity guidance network to weight features for the foreground predictions in the query branch. There are some previous works using recurrent structures to refine the segmentation predictions~\cite{hu2019attention,zhang2019canet}.
All previous methods only use the foreground mask in the query image as the training supervision, while in our network, the query set and the support set guide each other and both branches make foreground predictions for training supervision.

\subsection{Image co-segmentation}
Image co-segmentation is a well-studied task which aims to jointly segment the common objects in paired images. Many approaches have been proposed to solve the object co-segmentation problem.  Rotheret~\etal~\cite{rother2006cosegmentation} propose to minimize an energy function of a histogram matching term with an MRF to enforce similar foreground statistics. Rubinsteinet~\etal~\cite{rubinstein2013unsupervised} capture the sparsity and visual variability of  the common object from pairs of images with dense correspondences. Joulin ~\etal~\cite{joulin2012multi} solve the common object problem with an efficient convex quadratic approximation of energy with discriminate clustering. 
Since the prevalence of deep neural networks, many deep learning-based methods have been proposed. In~\cite{mukherjee2018object}, the model retrieves common object proposals with a Siamese network. Chen~\etal~ \cite{chen2018semantic} adopt channel attentions to weight features for the co-segmentation task. Deep learning-based approaches have significantly outperformed non-learning based methods.

\section{Task Definition} \label{problem_define}
Few-shot segmentation aims to find the foreground pixels in the test images given only a few pixel-level annotated images. The training and testing of the model are conducted on two datasets with no overlapped categories. At both the training and testing stages, the labeled example images are called the support set, which serves as a meta-training set and the unlabeled meta-testing image is called the query set. To guarantee a good generalization performance at test time, the training and evaluation of the model are accomplished by episodically sampling the support set and the query set.  

Given a network $\mathcal{R}_{\theta}$ parameterized by ${\theta}$, in each episode, we first sample a target category $c$ from the dataset $\mathcal{C}$. Based on the sampled class, we then sample $k+1$ labeled images $\{(x_s^1,y_s^1),(x_s^2,y_s^2),...(x_s^k,y_s^k),(x_q,y_q) \}$ that all contain the sampled category $c$. Among them, the first $k$ labeled images constitute the support set $\mathcal{S}$ and the last one is the query set $\mathcal{Q}$. After that, we make predictions on the query images by inputting the support set and the query image into the model $\hat y_{q}=\mathcal{R}_{\theta}(\mathcal{S},x_q)$. At training time, we learn the model parameters $\theta$ by optimizing the cross-entropy loss $\mathcal{L}(\hat y_{q},y_{q})$, and repeat such procedures until convergence.

\section{Method}

 \begin{figure*}[t]
    \includegraphics[width=1\linewidth]{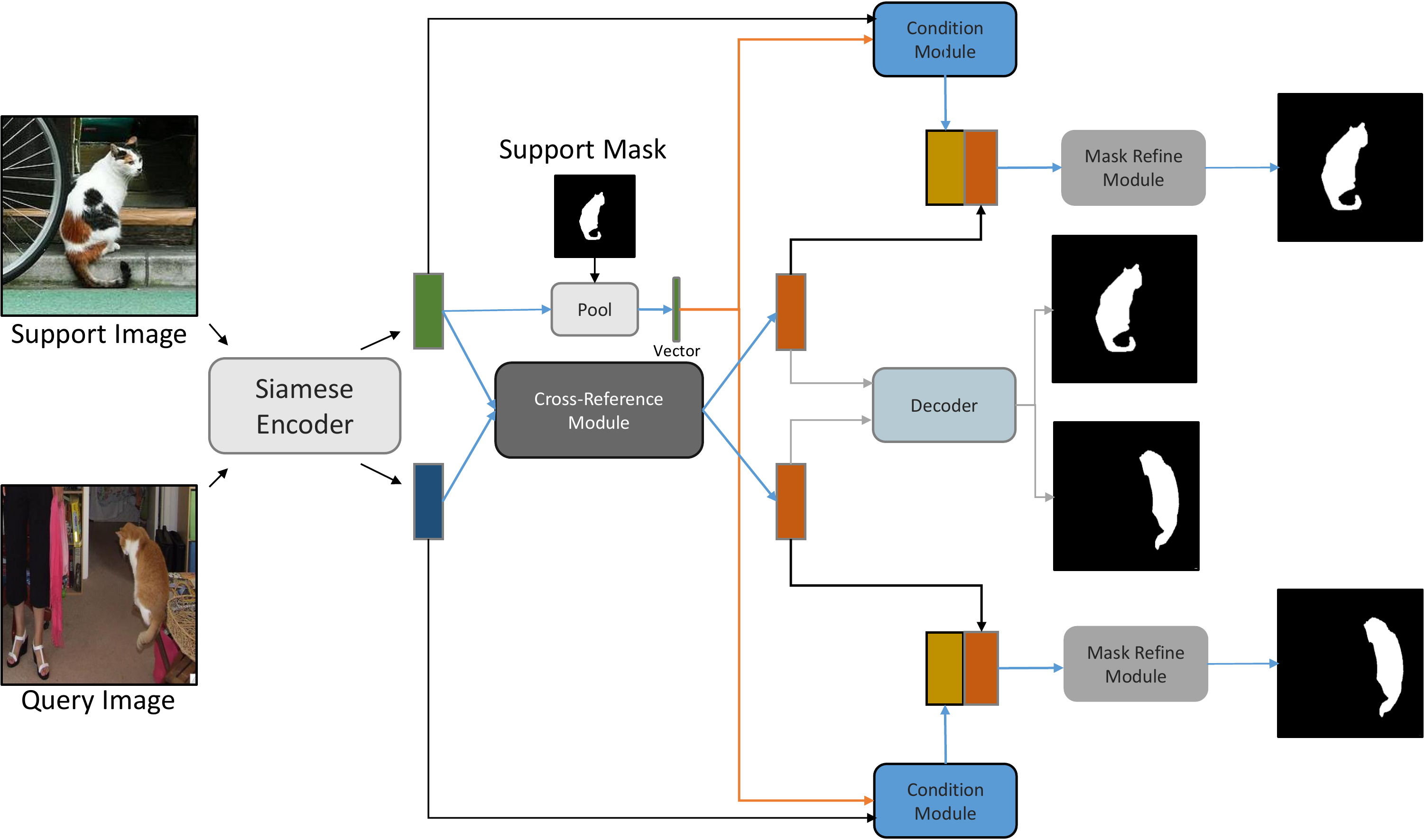}
    \vskip +1em
    \caption{The pipeline of our Network architecture. Our Network mainly consists of a Siamese encoder, a cross-reference module, a condition module, and a mask refinement module. Our network adopts a symmetric design. The Siamese encoder maps the query and support images into feature representations. The cross-reference module mines the co-occurrent features in two images to generate reinforced representations. The condition module fuses the category-relevant feature vectors into feature maps to emphasize the target category. The mask refinement module saves the confidence maps of the last prediction into the cache and recurrently refines the predicted masks.}
        \vskip +1em
    \label{archtecture}
\end{figure*}

In this section, we introduce the proposed cross-reference network for solving few-shot image segmentation. In the beginning, we describe our network in the 1-shot case. After that, we describe our finetuning scheme in the case of $k$-shot learning. Our network includes four key modules: the Siamese encoder, the cross-reference module, the condition module, and the mask refinement module. The overall architecture is shown in Fig.~\ref{archtecture}.

\subsection{Method overview}

Different from previous existing few-shot segmentation methods \cite{zhang2019canet,shaban2017one,Dong2018FewShotSS} unilaterally guide the segmentation of query images with support images, our proposed CRNet enables support and query images guide the segmentation of each other. 
We argue that the relationship between support-query image pairs is vital to few-shot segmentation learning. Experiments in Table~\ref{onefiveten} validate the effectiveness of our new architecture design.
As shown in Figure~\ref{archtecture}, our model learns to perform few-shot segmentation as follows: for every query-support pair, we encoder the image pair into deep features with the Siamese Encoder, then apply the cross-reference module to mine out co-occurrent object features. To fully utilize the annotated mask, the conditional module will incorporate the category information of support set annotations for foreground mask predictions, our mask refine module caches the confidence region maps recurrently for final foreground prediction. 
In the case of $k$-shot learning, previous works \cite{zhang2018sg,zhang2019canet,shaban2017one} on simply average the results of different 1-shot predictions, while we adopt an optimization-based method that finetunes the model to make use of more support data. Table~\ref{ablation:Fuse-and-FT} demonstrates the advantages of our method over previous works.

\subsection{Siamese encoder}
The Siamese encoder is a pair of parameter-shared convolutional neural networks that encode the query image and the support image to feature maps. Unlike the models in~\cite{shaban2017one,rakelly2018conditional}, we use a shared feature encoder to encode the support and the query images. By embedding the images into the same space, our cross-reference module can better mine co-occurrent features to locate the foreground regions. To acquire representative feature embeddings, we use skip connections to utilize multiple-layer features. As is observed in CNN feature visualization literature\cite{zhang2019canet,yosinski2015understanding}, features in lower layers often relate to low level cue and higher layers often relate to segment cue, we combine the lower level features and higher level features and passing to followed modules.

\subsection{Cross-Reference Module}
\begin{figure}[t]
    \includegraphics[width=1\linewidth]{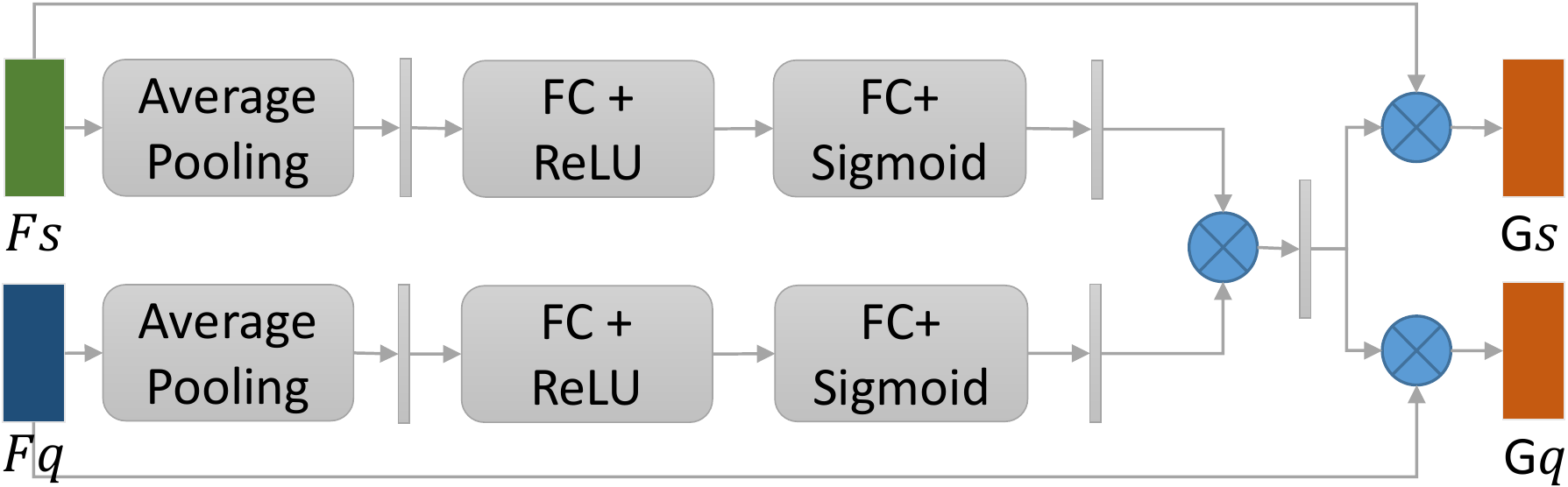}
    \caption{The cross-reference module. Given the input feature maps from the support and the query sets ($F_s$,$F_q$), the cross-reference module generates updated feature representations ($G_s$,$G_q$) by inspecting the co-occurrent features.}
    \label{co-segmentation}
\end{figure}

The cross-reference module is designed to mine co-occurrent features in two images and generate updated representations. The design of the module is shown in Fig.~\ref{co-segmentation}. Given two input feature maps generated by the Siamese encoder, we first use global average pooling to acquire the global statistics in the two images. Then, the two feature vectors are sent to a pair of two-layer fully connected (FC) layers, respectively. The Sigmoid activation function attached after the FC layer transforms the vector values into the importance of the channel, which is in the range of [0,1]. After that, the vectors in the two branches are fused by element-wise multiplication. Intuitively, only the common features in the two branches will have a high activation in the fused importance vector. Finally, we use the fused vector to weight the input feature maps to generate reinforced feature representations. In comparison to the raw features, the reinforced features focus more on the co-occurrent representations.  

Based on the reinforced feature representations, we add a head to directly predict the co-occurrent objects in the two images during training time.  This sub-task aims to facilitate the learning of the co-segmentation module to mine better feature representations for the downstream tasks. To generate the predictions of the co-occurrent objects in two images, the reinforced feature maps in the two branches are sent to a decoder to generate the predicted maps. 
The decoder is composed of convolutional layer followed by a ASPP\cite{chen2018deeplab} layers, finally, a convolutional layer generates a two-channel prediction corresponding to the foreground and background scores. 

\subsection{Condition Module}

\begin{figure}[t]
    \includegraphics[width=1\linewidth]{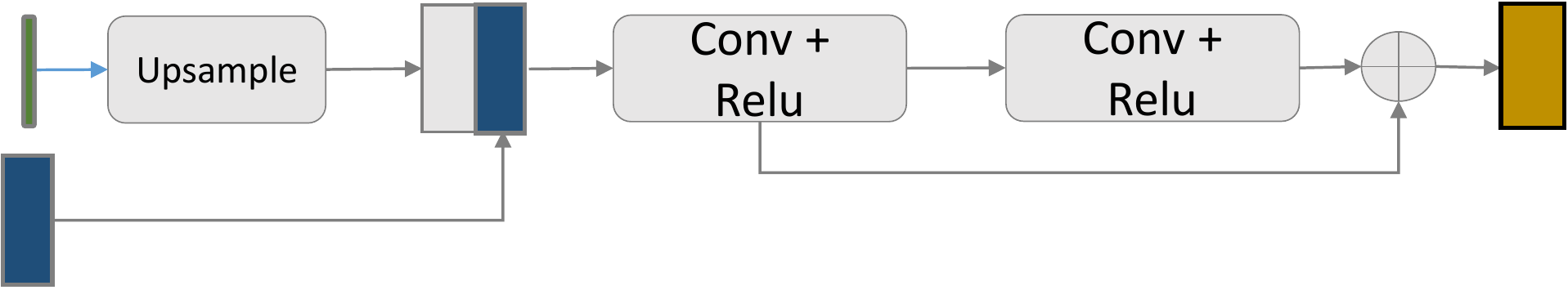}
    \caption{The condition module. Our condition module fuses the category-relevant features into representations for better predictions of the target category.}
    \label{refernet}
\end{figure}

To fully utilize the support set annotations, we design a condition module to efficiently incorporate the category information for foreground mask predictions. The condition module takes the reinforced feature representations generated by the cross-reference module and a category-relevant vector as inputs. The category-relevant vector is the fused feature embeddings of the target category, which is achieved by applying foreground average pooling\cite{zhang2019canet} over the category region. As the goal of the few-shot segmentation is to only find the foreground mask of the assigned object category, the task-relevant vector serves as a condition to segment the target category. To achieve a category-relevant embedding, previous works opt to filter out the background regions in the input images~\cite{rakelly2018conditional,shaban2017one} or in the feature representations\cite{zhang2019canet,zhang2018sg}. We choose to do so both in the feature level and in the input image. The category-relevant vector is fused with the reinforced feature maps in the condition module by bilinearly upsampling the vector to the same spatial size of the feature maps and concatenating them. Finally, we add a residual convolution to process the concatenated features. The structure of the condition module can be found in Fig.~\ref{refernet}. The condition modules in the support branch and the query branch have the same structure and share all the parameters.

\subsection{Mask Refinement Module}
\begin{figure*}[t]
    \includegraphics[width=1\linewidth]{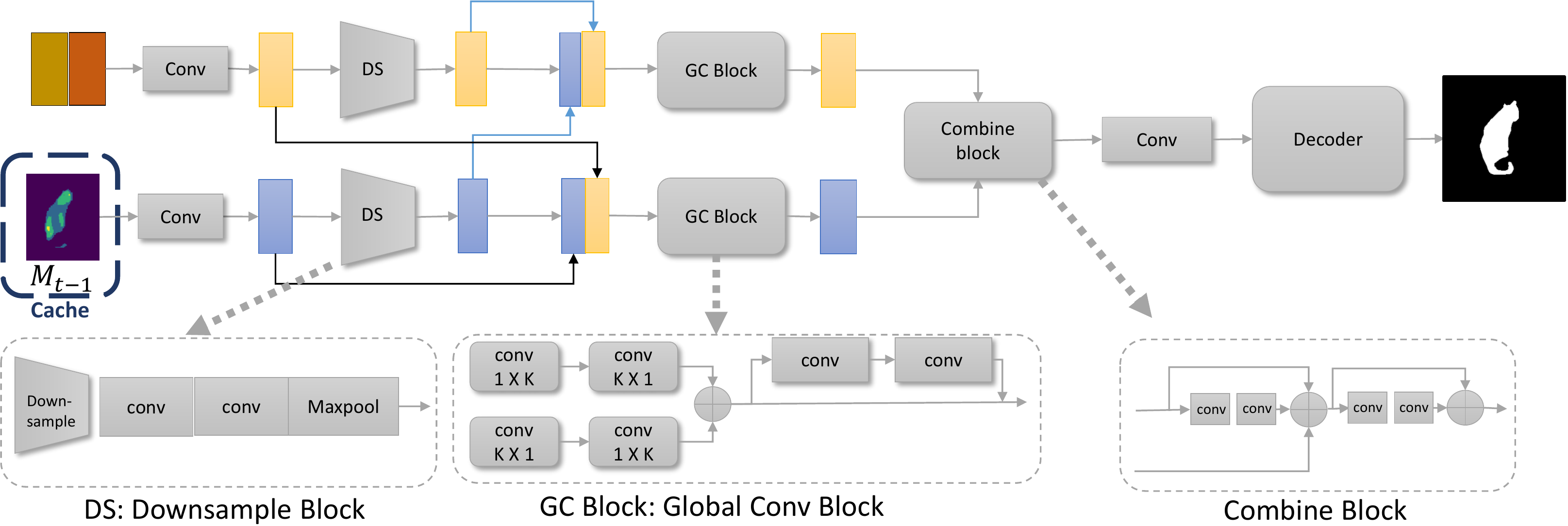}
    \caption{The mask refinement module. The module saves the generated probability map from the last step into the cache and recurrently optimizes the predictions. }
    \label{memory}
\end{figure*}

As is often observed in the weakly supervised semantic segmentation literature~\cite{zhang2019canet,kolesnikov2016seed}, directly predicting the object masks can be difficult. It is a common principle to firstly locate seed regions and then refine the results. Based on such principle, we design a mask refinement module to refine the predicted mask step-by-step. Our motivation is that the probability maps in a single feed-forward prediction can reflect where is the confident region in the model prediction. Based on the confident regions and the image features, we can gradually optimize the mask and find the whole object regions. As shown in Fig.~\ref{memory}, our mask refinement module has two inputs. One is the saved confidence map in the cache and the second input is the concatenation of the outputs from the condition module and the cross-reference module. For the initial prediction, the cache is initialized with a zero mask, and the module makes predictions solely based on the input feature maps. The module cache is updated with the generated probability map every time the module makes a new prediction. We run this module multiple times to generate a final refined mask.

The mask refinement module includes three main blocks: the downsample block, the global convolution block, and the combine block.
The Downsample Block downsamples the feature maps by a factor of 2. The downsampled features are then upsampled to the original size and fused with features in the opposite branch. The global convolution block~\cite{peng2017large} aims to capture features in a large field-of-view while containing few parameters. It includes two groups of $1\times7$ and $7\times1$ convolutional kernels. The combine block effectively fuses the feature branch and the cached branch to generate refined feature representations.

\subsection{Finetuning for K-Shot Learning}
In the case of $k$-shot learning, we propose to finetune our network to take advantage of multiple labeled support images. As our network can make predictions for two images at a time, we can use at most $k^2$ image pairs to finetune our network. At the evaluation stage, we randomly sample an image pair from the labeled support set to finetune our model. We keep the parameters in the Siamese encoder fixed and only finetune the rest modules. In our experiment, we demonstrate that our finetuning based methods can consistently improve the result when more labeled support images are available, while the fusion-based methods in previous works often get saturated performance when the number of support images increases.

\section{Experiment}

 \begin{figure*}[t]
    \includegraphics[width=1\linewidth,height=0.70\linewidth]{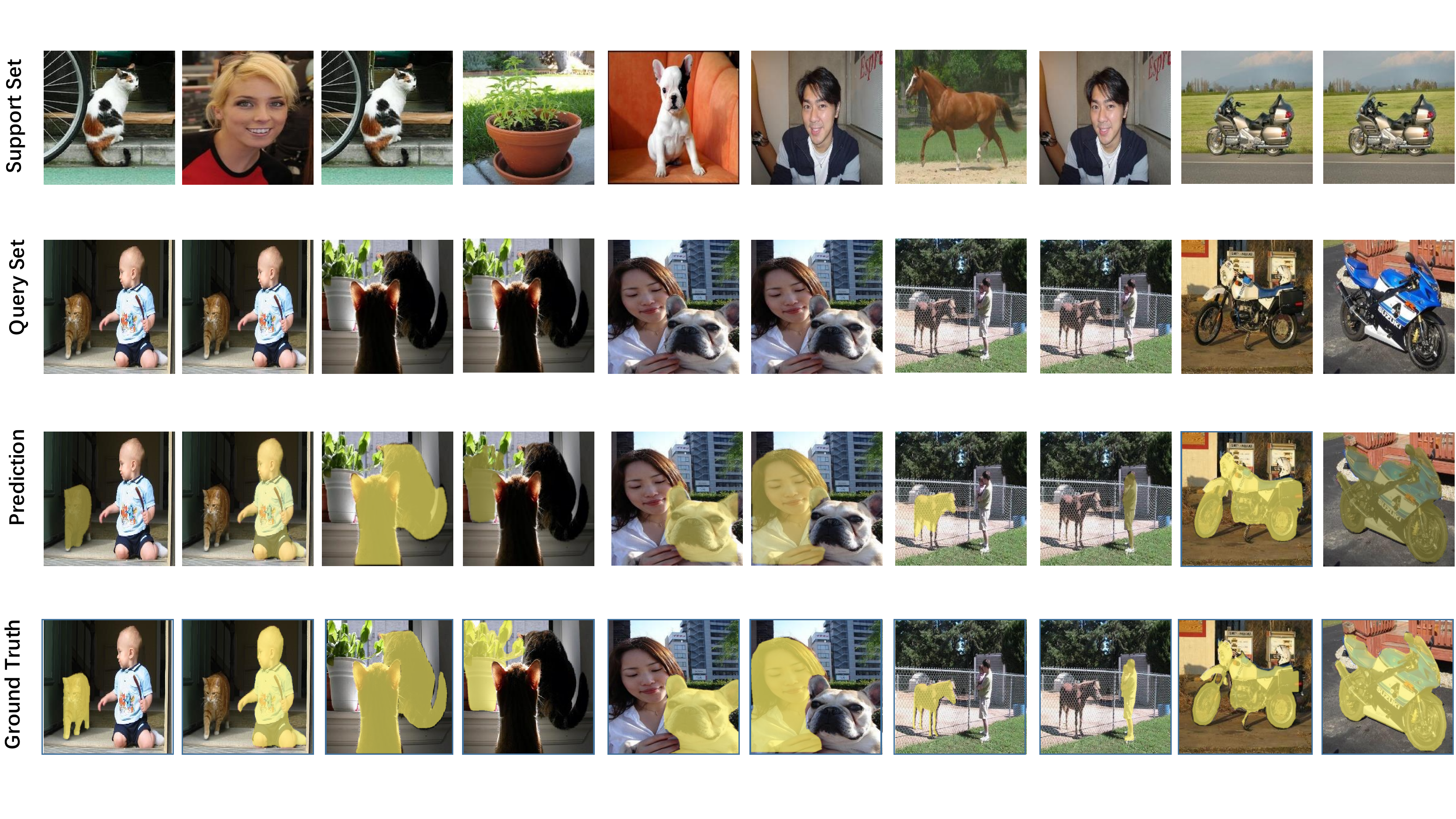}
    \caption{Our Qualitative examples on the PASCAL VOC dataset. The first row is the support set and the second row is the query set. The third row is our predicted results and the the fourth row is the ground truth. Even when the query images contain objects from multiple classes, our network can still successfully segment the target category indicated by the support mask. }
    \vskip +1em
    \label{results}
\end{figure*}

\subsection{Implementation Details}
In the Siamese encoder, we exploit multi-level features from the ImageNet pre-trained Resnet-50 as the image representations. We use dilated convolutions and keep the feature maps after layer3 and layer4 have a fixed size of 1/8 of the input image and concatenate them for final prediction.
All the convolutional layers in our proposed modules have the kernel size of $3\times3$ and generate features of 256 channels, followed by the ReLU activation function. 
At test time, we recurrently run the mask refinement module for 5 times to refine the predicted masks. In the case of $k$-shot learning, we fix the Siamese encoder and finetune the rest parameters.

\subsection{Dataset and Evaluation Metric}
We implement cross-validation experiments on the PASCAL VOC 2012 dataset to validate our network design. To compare our model with previous works, we adopt the same category divisions and test settings which are first proposed in~\cite{shaban2017one}. In the cross-validation experiments, 20 object categories are evenly divided into 4 folds, with three folds as the training classes and one fold as the testing classes. The category division is shown in Table~\ref{division}. We report the average performance over 4 testing folds. For the evaluation metrics, we use the standard mean Intersection-over-Union (mIoU) of the classes in the testing fold. For more details about the dataset information and the evaluation metric, please refer to~\cite{shaban2017one}.

\begin{table}[t]
\centering
\small
\resizebox{0.9\columnwidth}{!}
{
\begin{tabular}{c|c}
\toprule[1pt]
fold & categories \\\hline
0 & aeroplane, bicycle, bird, boat, bottle  \\ \hline
1 & bus, car, cat, chair, cow        \\\hline
2 & diningtable, dog, horse, motobike, person          \\\hline
3 &  potted plant, sheep, sofa, train, tv/monitor       \\
\bottomrule[1pt]
\end{tabular}%
}
        \vskip +1em
\caption{The class division of the  PASCAL VOC 2012 dataset proposed in~\cite{shaban2017one}. }
\label{division}
\end{table}

\begin{table}[t]
\centering
\small
\resizebox{0.8\columnwidth}{!}
{

\begin{tabular}{ccccc}
\toprule[1pt]
Condition & Cross-Reference Module  & 1-shot  \\
\hline
\checkmark   &        & 36.3       \\
    & \checkmark      & 43.3         \\
\checkmark   & \checkmark      & 49.1    \\ 
\bottomrule[1pt]
\end{tabular}

}
        \vskip +1em
\caption{Ablation study on the condition module and the cross-reference module. The cross-reference module brings a large performance improvement over the baseline model (Condition only).}

\label{onefiveten}
\end{table}

\begin{table}[t]
\centering
\small
\resizebox{0.8\columnwidth}{!}
{

\begin{tabular}{lllll}
\toprule[1pt]
Multi-Level & Mask Refine & Multi-Scale & 1-shot  \\
\hline
  &    &    & 49.1      \\
  &    & \checkmark  & 50.3       \\
  & \checkmark  & \checkmark  & 53.4     \\
\checkmark  & \checkmark  & \checkmark  & 55.2   \\
\bottomrule[1pt]
\end{tabular}

}
        \vskip +1em
\caption{Ablation experiments on the multiple-level feature, multiple-scale input, and the Mask Refine module. Every module brings performance improvement over the baseline model.
}
\label{table:different-tech}
\end{table}

\begin{table}[t]
\centering
\small
\resizebox{0.9\columnwidth}{!}
{

\begin{tabular}{llll}
\toprule[1pt]
Method & 1-shot & 5-shot & 10-shot \\
\hline
Fusion   & 49.1   & 50.2   & 49.9    \\
Finetune & N/A   & 57.5   & 59.1  \\
Finetune + Fusion &N/A & 57.6   & 58.8  \\
\bottomrule[1pt]
\end{tabular}
}

        \vskip +1em
\caption{$k$-shot experiments. We compare our finetuning based method with the fusion method. Our method yields consistent performance improvement when the number of support images increases. For the case of $1$-shot, finetune results are not available as CRNet needs at least two images to apply our finetune scheme. 
}
\label{ablation:Fuse-and-FT}
\end{table}

\begin{table}[t]
\centering
\small
\resizebox{0.8\columnwidth}{!}
{

\begin{tabular}{lccc}
\toprule[1pt]
Method       & Backbone &  mIoU &  IoU \\
\hline
OSLM\cite{shaban2017one}         & VGG16      & 40.8        & 61.3          \\
co-fcn\cite{rakelly2018conditional}       & VGG16      & 41.1        & 60.9          \\
sg-one \cite{zhang2018sg}      & VGG16      & 46.3        & 63.1          \\
R-DRCN \cite{siam2019adaptive} & VGG16      & 40.1        & 60.9          \\
PL\cite{Dong2018FewShotSS}           & VGG16      & -           & 61.2          \\
A-MCG\cite{hu2019attention}        & ResNet-50   & -           & 61.2          \\
CANet\cite{zhang2019canet}        & ResNet-50   & 55.4    & 66.2         \\
PGNet\cite{zhangchi2} & ResNet-50   & \textbf{56.0}      & \textbf{69.9}         \\
\hline
CRNet         & VGG16      & 55.2        & 66.4     \\
CRNet       &ResNet-50      & 55.7      & 66.8 \\
\bottomrule[1pt]
\end{tabular}
}

\vskip +1em
\caption{Comparison with the state-of-the-art methods under the 1-shot setting. Our proposed network achieves state-of-the-art performance under both evaluation metrics. }
\label{table:1-shot-iou}
\end{table}

\begin{table}[t]
\centering
\small
\resizebox{0.8\columnwidth}{!}
{
\begin{tabular}{lccc}
\toprule[1pt]
Method       & Backbone &  mIoU & IoU \\
\hline
OSLM \cite{shaban2017one}        & VGG16      & 43.9        & 61.5          \\
co-fcn \cite{rakelly2018conditional}      & VGG16      & 41.4        & 60.2          \\
sg-one  \cite{zhang2018sg}     & VGG16      & 47.1        & 65.9          \\
R-DFCN \cite{siam2019adaptive} & VGG16      & 45.3        & 66.0          \\
PL   \cite{Dong2018FewShotSS}        & VGG16      & -           & 62.3          \\
A-MCG    \cite{hu2019attention}    & ResNet-50   & -           & 62.2          \\
CANet \cite{zhang2019canet}       & ResNet-50   & 57.1         & 69.6          \\

PGNet\cite{zhangchi2} &ResNet50     &58.5           &70.5 \\
\hline
CRNet & VGG16      & 58.5      & 71.0 \\
CRNet         & ResNet50      & \textbf{58.8}        & \textbf{71.5}     \\
\bottomrule[1pt]
\end{tabular}
}
\vskip +1em
\caption{Comparison with the state-of-the-art methods under the 5-shot setting. Our proposed network outperforms all previous methods and achieves new state-of-the-art performance under both evaluation metrics. }
\label{table-5-shot-iou}
\end{table}

\section{Ablation study} \label{ablationsec}
The goal of the ablation study is to inspect each component in our network design. 
Our ablation experiments are conducted on the PASCAL VOC dataset. We implement cross-validation 1-shot experiments and report the average performance over the four splits.

In Table~\ref{onefiveten}, we first investigate the contributions of our two important network components: the condition module and the cross-reference module. As shown, there are significant performance drops if we remove either component from the network. Particularly, our proposed cross-reference module has a huge impact on the predictions. Our network can improve the counterpart model without cross-reference module by more than 10\%.

To investigate how much the scale variance of the objects influence the network performance, we adopt a multi-scale test experiment in our network. Specifically, at the test time, we resize the support image and the query image to [0.75,1.25] of the original image size and conduct the inference. The output predicted mask of the resized query image is bilinearly resized to the original image size. We fuse the predictions under different image scales. As shown in Table~\ref{table:different-tech}, multi-scale input test brings 1.2 mIoU score  improvement in the 1-shot setting. 
We also investigate the choices of features in the network backbone in Table~\ref{table:different-tech}. We compare the multi-level feature embeddings with the features solely from the last layer. Our model with multi-level features provides an improvement of 1.8 mIoU score. This indicates that to better locate the common objects in two images, middle-level features are also important and helpful. 

To further inspect the effectiveness
of the mask refinement module, we design a baseline model
that removes the cached branch. In this case, the mask refinement block makes predictions solely based on the input
features and we only run the mask refinement module once.
As shown in Table~\ref{table:different-tech} , our mask refinement module brings
3.1 mIoU score performance increase over our baseline method.

In the $k$-shot setting, we compare our finetuning based method with the fusion-based methods widely used in previous works. For the fusion-based method, we make an inference with each of the support images and average their probability maps as the final prediction. The comparison is shown in Table~\ref{ablation:Fuse-and-FT}. In the 5-shot setting, the finetuning based method  outperforms 1-shot baseline by 8.4 mIoU score, which is significantly superior to the fusion-based method. When 10 support images are available, our finetuning based method shows more advantages. The performance continues increasing while the fusion-based method's performance begins to drop.

\subsection{MS COCO}
COCO 2014~\cite{coco} is a challenging large-scale dataset, which contains 80 object categories.
Following~\cite{zhang2019canet}, we choose 40 classes for training, 20 classes for validation and 20 classes for test.
As shown in Table.~\ref{coco-data}, the results again validate the designs in our network.

\subsection{Comparison with the State-of-the-Art Results}
We compare our network with state-of-the-art methods on the PASCAL VOC 2012 dataset. Table~\ref{table:1-shot-iou} shows the performance of different methods in the 1-shot setting. We use IoU to denote the evaluation metric proposed in~\cite{rakelly2018conditional}. The difference between the two metrics is that the IoU metric also incorporates the background into the Intersection-over-Union computation and ignores the image category. 

\textbf{5-Shot Experiments.} The comparison of 5-shot segmentation results under two evaluation metrics is shown in Table~\ref{table-5-shot-iou}. Our method achieves new state-of-the-art performance under both evaluation metrics.

\begin{table}[t]
\centering
\small
\resizebox{0.9\columnwidth}{!}
{

\begin{tabular}{ccccc}
\toprule[1pt]
Condition & Cross-Reference Module & Mask-Refine & 1-shot & 5-shot \\
\hline
\checkmark   &               &   & 43.3  & 44.0      \\
              & \checkmark   &   & 38.5  & 42.7      \\
\checkmark   & \checkmark    &   & 44.9  & 45.6       \\ 
\checkmark   & \checkmark    &  \checkmark  & 45.8  & 47.2  \\ 
\bottomrule[1pt]
\end{tabular}

}
        \vskip +1em

\caption{Ablation study on the condition module cross-reference module and Mask-refine module on dataset MS COCO.}

\label{coco-data}
\end{table}

\section{Conclusion}
In this paper, we have presented a novel cross-reference network for few-shot segmentation. Unlike previous work unilaterally guiding the segmentation of query images with support images, our two-head design concurrently makes predictions in both the query image and the support image to help the network better locate the target category. We develop a mask refinement module with a cache mechanism which can effectively improve the prediction performance. In the $k$-shot setting, our finetuning based method can take advantage of more annotated data and significantly improves the performance. Extensive ablation experiments on PASCAL VOC 2012 dataset validate the effectiveness of our design. Our model achieves state-of-the-art performance on the PASCAL VOC 2012 dataset.
\section*{Acknowledgements} This research is supported by the National Research Foundation Singapore under its AI Singapore Programme (Award Number: AISG-RP-2018-003) and the MOE Tier-1 research grants: RG126/17 (S) and RG22/19 (S). This research is also partly supported by the Delta-NTU Corporate Lab with funding support from Delta Electronics Inc. and the National Research Foundation (NRF) Singapore.

{\small
\bibliographystyle{ieee_fullname}
\bibliography{egbib}

\begin{thebibliography}{10}\itemsep=-1pt

\bibitem{chen2018semantic}
Hong Chen, Yifei Huang, and Hideki Nakayama.
\newblock Semantic aware attention based deep object co-segmentation.
\newblock {\em arXiv preprint arXiv:1810.06859}, 2018.

\bibitem{chen2018deeplab}
Liang-Chieh Chen, George Papandreou, Iasonas Kokkinos, Kevin Murphy, and Alan~L
  Yuille.
\newblock Deeplab: Semantic image segmentation with deep convolutional nets,
  atrous convolution, and fully connected crfs.
\newblock {\em IEEE transactions on pattern analysis and machine intelligence},
  40(4):834--848, 2018.

\bibitem{chen2019closerfewshot}
Wei-Yu Chen, Yen-Cheng Liu, Zsolt Kira, Yu-Chiang Wang, and Jia-Bin Huang.
\newblock A closer look at few-shot classification.
\newblock In {\em International Conference on Learning Representations}, 2019.

\bibitem{imagenet}
Jia Deng, Wei Dong, Richard Socher, Li-Jia Li, Kai Li, and Li Fei-Fei.
\newblock Imagenet: A large-scale hierarchical image database.
\newblock In {\em CVPR}, pages 248--255, 2009.

\bibitem{Dong2018FewShotSS}
Nanqing Dong and Eric Xing.
\newblock Few-shot semantic segmentation with prototype learning.
\newblock In {\em BMVC}, 2018.

\bibitem{hu2019attention}
Tao Hu, Pengwan Yang, Chiliang Zhang, Gang Yu, Yadong Mu, and Cees~GM Snoek.
\newblock Attention-based multi-context guiding for few-shot semantic
  segmentation.
\newblock 2019.

\bibitem{joulin2012multi}
Armand Joulin, Francis Bach, and Jean Ponce.
\newblock Multi-class cosegmentation.
\newblock In {\em 2012 IEEE Conference on Computer Vision and Pattern
  Recognition}, pages 542--549. IEEE, 2012.

\bibitem{kolesnikov2016seed}
Alexander Kolesnikov and Christoph~H Lampert.
\newblock Seed, expand and constrain: Three principles for weakly-supervised
  image segmentation.
\newblock In {\em European Conference on Computer Vision}, pages 695--711.
  Springer, 2016.

\bibitem{lin2017refinenet}
Guosheng Lin, Anton Milan, Chunhua Shen, and Ian~D Reid.
\newblock Refinenet: Multi-path refinement networks for high-resolution
  semantic segmentation.
\newblock In {\em CVPR}, volume~1, page~5, 2017.

\bibitem{coco}
Tsung-Yi Lin, Michael Maire, Serge Belongie, James Hays, Pietro Perona, Deva
  Ramanan, Piotr Doll{\'a}r, and C~Lawrence Zitnick.
\newblock Microsoft coco: Common objects in context.
\newblock In {\em ECCV}, pages 740--755, 2014.

\bibitem{long2015fully}
Jonathan Long, Evan Shelhamer, and Trevor Darrell.
\newblock Fully convolutional networks for semantic segmentation.
\newblock In {\em Proceedings of the IEEE conference on computer vision and
  pattern recognition}, pages 3431--3440, 2015.

\bibitem{mukherjee2018object}
Prerana Mukherjee, Brejesh Lall, and Snehith Lattupally.
\newblock Object cosegmentation using deep siamese network.
\newblock {\em arXiv preprint arXiv:1803.02555}, 2018.

\bibitem{peng2017large}
Chao Peng, Xiangyu Zhang, Gang Yu, Guiming Luo, and Jian Sun.
\newblock Large kernel matters--improve semantic segmentation by global
  convolutional network.
\newblock In {\em Proceedings of the IEEE conference on computer vision and
  pattern recognition}, pages 4353--4361, 2017.

\bibitem{rakelly2018conditional}
Kate Rakelly, Evan Shelhamer, Trevor Darrell, Alyosha Efros, and Sergey Levine.
\newblock Conditional networks for few-shot semantic segmentation.
\newblock In {\em ICLR Workshop}, 2018.

\bibitem{rother2006cosegmentation}
Carsten Rother, Tom Minka, Andrew Blake, and Vladimir Kolmogorov.
\newblock Cosegmentation of image pairs by histogram matching-incorporating a
  global constraint into mrfs.
\newblock In {\em 2006 IEEE Computer Society Conference on Computer Vision and
  Pattern Recognition (CVPR'06)}, volume~1, pages 993--1000. IEEE, 2006.

\bibitem{rubinstein2013unsupervised}
Michael Rubinstein, Armand Joulin, Johannes Kopf, and Ce Liu.
\newblock Unsupervised joint object discovery and segmentation in internet
  images.
\newblock In {\em Proceedings of the IEEE conference on computer vision and
  pattern recognition}, pages 1939--1946, 2013.

\bibitem{shaban2017one}
Amirreza Shaban, Shray Bansal, Zhen Liu, Irfan Essa, and Byron Boots.
\newblock One-shot learning for semantic segmentation.
\newblock {\em arXiv preprint arXiv:1709.03410}, 2017.

\bibitem{siam2019adaptive}
Mennatullah Siam and Boris Oreshkin.
\newblock Adaptive masked weight imprinting for few-shot segmentation.
\newblock {\em arXiv preprint arXiv:1902.11123}, 2019.

\bibitem{snell2017prototypical}
Jake Snell, Kevin Swersky, and Richard Zemel.
\newblock Prototypical networks for few-shot learning.
\newblock In {\em NIPS}, 2017.

\bibitem{sun2020conditional}
Xin Sun, Zhenning Yang, Chi Zhang, Guohao Peng, and Keck-Voon Ling.
\newblock Conditional gaussian distribution learning for open set recognition,
  2020.

\bibitem{vinyals2016matching}
Oriol Vinyals, Charles Blundell, Timothy Lillicrap, Daan Wierstra, et~al.
\newblock Matching networks for one shot learning.
\newblock In {\em Advances in neural information processing systems}, pages
  3630--3638, 2016.

\bibitem{relation}
Flood Sung~Yongxin Yang, Li Zhang, Tao Xiang, Philip~HS Torr, and Timothy~M
  Hospedales.
\newblock Learning to compare: Relation network for few-shot learning.
\newblock In {\em CVPR}, 2018.

\bibitem{yosinski2015understanding}
Jason Yosinski, Jeff Clune, Anh Nguyen, Thomas Fuchs, and Hod Lipson.
\newblock Understanding neural networks through deep visualization.
\newblock {\em arXiv preprint arXiv:1506.06579}, 2015.

\bibitem{zhangcvpr}
Chi Zhang, Yujun Cai, Guosheng Lin, and Chunhua Shen.
\newblock Deepemd: Few-shot image classification with differentiable earth
  mover's distance and structured classifiers, 2020.

\bibitem{zhangchi2}
Chi Zhang, Guosheng Lin, Fayao Liu, Jiushuang Guo, Qingyao Wu, and Rui Yao.
\newblock Pyramid graph networks with connection attentions for region-based
  one-shot semantic segmentation.
\newblock In {\em Proceedings of the IEEE International Conference on Computer
  Vision}, pages 9587--9595, 2019.

\bibitem{zhang2019canet}
Chi Zhang, Guosheng Lin, Fayao Liu, Rui Yao, and Chunhua Shen.
\newblock Canet: Class-agnostic segmentation networks with iterative refinement
  and attentive few-shot learning.
\newblock In {\em Proceedings of the IEEE Conference on Computer Vision and
  Pattern Recognition}, pages 5217--5226, 2019.

\bibitem{zhang2018sg}
Xiaolin Zhang, Yunchao Wei, Yi Yang, and Thomas Huang.
\newblock Sg-one: Similarity guidance network for one-shot semantic
  segmentation.
\newblock {\em arXiv preprint arXiv:1810.09091}, 2018.

\end{thebibliography}
}
\end{document}